\documentclass[letterpaper]{article}
\usepackage{aaai}
\usepackage{graphicx}
\usepackage{color,soul}
\usepackage{amsmath, amsfonts, amssymb}
\usepackage[ruled,vlined,linesnumbered]{algorithm2e}
\usepackage{comment}
\usepackage{subcaption}
\usepackage{times}
\usepackage{helvet}  
\usepackage{courier}
\usepackage{array}
\newcolumntype{L}[1]{>{\raggedright\let\newline\\\arraybackslash\hspace{0pt}}m{#1}}
\newcolumntype{C}[1]{>{\centering\let\newline\\\arraybackslash\hspace{0pt}}m{#1}}
\newcolumntype{R}[1]{>{\raggedleft\let\newline\\\arraybackslash\hspace{0pt}}m{#1}}
\begin{document}
%
\title{Online Article Ranking as a Constrained, Dynamic, \\Multi-Objective Optimization Problem}

\author{Jeya Balaji Balasubramanian \\ Intelligent Systems Program, \\ University of Pittsburgh, Pittsburgh \\ jeya@pitt.edu \And Akshay Soni \\ Yahoo! Research, Sunnyvale \\ akshaysoni@yahoo-inc.com \AND Yashar Mehdad \\ Airbnb, San Francisco \\yashar.mehdad@airbnb.com \And Nikolay Laptev \\ University of California, Los Angeles \\nlaptev@ucla.edu}
\maketitle
\begin{abstract}
\begin{quote}
The content ranking problem in a social news website, is typically a function that maximizes a scalar metric of interest like dwell-time. However, like in most real-world applications we are interested in more than one metric---for instance simultaneously maximizing click-through rate, monetization metrics, dwell-time---and also satisfy the traffic requirements promised to different publishers. All this needs to be done on online data and under the settings where the objective function and the constraints can dynamically change; this could happen if for instance new publishers are added, some contracts are adjusted, or if some contracts are over. 

In this paper, we formulate this problem as a \emph{constrained}, \emph{dynamic}, \emph{multi-objective} optimization problem. We propose a novel framework that extends a successful genetic optimization algorithm, NSGA-II, to solve this online, data-driven problem. We design the modules of NSGA-II to suit our problem. We evaluate optimization performance using Hypervolume and introduce a confidence interval metric for assessing the practicality of a solution. We demonstrate the application of this framework on a real-world Article Ranking problem. We observe that we make considerable improvements in both time and performance over a brute-force baseline technique that is currently in production.
\end{quote}
\end{abstract}

\section{Introduction}

Ranking a list of content based on a query (e.g., search results, or news articles where the query is the user) has been studied in depth. An example of this is article ranking in a news website that is typically aimed at maximizing a single scalar objective like dwell-time or click-through-rate (CTR). These are known as single-objective optimization (SOO) problems. However, most real-world content ranking platforms tries to optimize multiple objectives like dwell-time, CTR, daily active users (DAU), and monetization metrics simultaneously. Note that some of these objectives may be conflicting in the sense that if one of them increases, others may be forced to decrease; for instance, generally if one tries to increase the dwell-time, then CTR drops. This generates the need for formulating this as a multi-objective optimization (MOO) problem.

To make it further complicated, the article ranking problems are generally subject to constraints such as \emph{traffic shaping} requirements---these are partnership contracts between content aggregators and third party content providers to acquire content in return for a target traffic or revenue promises. The traffic, or revenue commitment, is documented in a contract and then manually tracked and reported by the partnership lead. This is usually done by pulling the data from various tools and sending a spreadsheet to the partners. This is a highly laborious process that is not scalable while also not providing the level of information granularity needed. One major challenge of many content aggregators is the control over achieving these targets; the current model is based on after-the-fact reporting rather than controlled targeting. The gap between the commitment and actual target achievement is often large and can range from 10\% to 50\% under achievement.

Finally, the objectives of interest are generally dynamic in nature i.e., they may change with time. For example, CTR is known to show significant temporal variations with time-of-day, day-of-week effect, and CTR decay due to repeated exposure to the same articles \cite{agarwal2009spatio}. This requires us to optimize target traffic with respect to the dynamic behavior of these objectives.

\subsection{Problem description}
An \textit{objective variable}, $y$, is any random variable of interest that we want to optimize. An \textit{objective space}, $Y$, is a vector of all objective variables that we want to simultaneously optimize. We represent the objective space as an $m$-dimensional vector, $\vec{y} = ( y_1, y_2,\dots,y_m )$ with $m$ different objectives. For example in the news ranking problem, we may want to simultaneously optimize on total clicks, and dwell-time. A \textit{design variable}, $x$, is any other random variable that can affect the value of the objective variables. This is usually a variable we can easily measure and control. The \textit{design space}, $X$, is a vector of all design variables that can affect our objectives. We represent the design space as an $n$-dimensional vector, $\vec{x} = ( x_1, x_2,\dots,x_n )$, with $n$ candidate design variables. Examples of design variables in ranking can include the amount of user interest (views, likes, comments), page presentation, publisher reputation, etc. An \textit{objective function} is a mapping function from the design space to the objective space, $f:X \rightarrow Y$. A multi-objective function can be represented as a vector of $m$ scalar objective functions, one for each objective variable, $F(\vec{x}) = (f_1(\vec{x}), f_2(\vec{x}), \dots, f_m(\vec{x}))$.

We formulate the generic MOO problem as the following minimization problem,
\begin{align}
    & \min_{\vec{x} \in S} F(\vec{x}) = [f_{1}(\vec{x}), f_{2}(\vec{x}),\dots ,f_{m}(\vec{x})]\\
    & \text{where, } S = \{\vec{x} \in \mathbb{R}^{n}: h(\vec{x}) = 0, g(\vec{x}) \geq 0\}\nonumber
\end{align}
The objective space can be subject to a set of constraints specified by the set $S$. The constraints can be represented in terms of equality constraints, $h(\cdot)$, and/or inequality constraints, $g(\cdot)$.  For example, in traffic shaping, these constraints can be a requirement for total clicks to be greater than some target value for a partner, $\theta_{clicks}$. 
This constraint can be re-written as: $f_{Clicks}(\vec{x}) - \theta_{Clicks} \geq 0$ or $g(\vec{x}) \geq 0$, where $g(\vec{x}) = f_{Clicks}(\vec{x}) - \theta_{Clicks}$.

The image of the feasible set under $F(\vec{x})$ for all possible values of $\vec{x}$ under constraints $S$ is called the \textit{attained set}. This is represented as $C = \{\vec{y} \in \mathbb{R}^{m}: \vec{y}=F(\vec{x}), \vec{x} \in S\}$. Typically, a single-objective optimization returns a single globally optimal solution. In contrast, a solution to a MOO problem is a set of solutions, where each solution is said to be \textit{Pareto optimal}.

\textbf{Definition 1.} \textit{Pareto optimal: A solution $\vec{x}^* \in X$ to a minimization problem is said to be Pareto optimal iff there does not exist another solution $\vec{x} \in X$, such that, $F(\vec{x}) \leq F(\vec{x}^*)$, and $f_i(\vec{x}) < f_i(\vec{x}^*)$ for at least one of the objectives.} \cite{marler2004survey}

A \textit{Pareto set} (or \textit{Pareto front}) is a set of all Pareto optimal solutions. This set is composed of extreme solutions (design variable assignments) which is optimal for one of the $m$ objectives, while it is sub-optimal on others. All other solutions represent a range of trade-offs between the different objectives. In practice, it is not feasible to compute all solutions in the attained set $C$, therefore we require approximate solutions with some theoretical guarantees.

There are two important properties, or theoretical guarantees, of a good solution set that we are interested in— 1) \textit{Convergence}: each solution should be as close to the global optimum as possible; and 2) \textit{Diversity}: the solution set should be uniformly spread over the Pareto front. Convergence ensures optimal solutions, and diversity ensures that we cover a wide variety of trade-offs between the multiple objectives.

Let variable $t$ represent time. The generalized dynamic MOO can be formulated using Equation (1) with a time parameter
\begin{align}
    & \min_{\vec{x} \in S_t} F(\vec{x},t) = [f_{1}(\vec{x},t), f_{2}(\vec{x},t),\dots ,f_{m}(\vec{x},t)]\\
    & \text{where, }S_t = \{\vec{x} \in \mathbb{R}^{n}: h(\vec{x},t) = 0, g(\vec{x},t) \geq 0\}\nonumber
\end{align}

Our contributions in this work include the development of a novel framework to perform constrained, MOO and extend it to optimize on dynamic objectives. To the best of our knowledge, this is the first real-world application of MOO in an online data-driven problem. We propose modules in our framework more suited to the online problem. This presented us with new challenges including practicality of the solutions. We address those concerns with a new metric called Confidence Uncertainty. 

\section{Approach}
We start by first discussing the current ranking approach for a popular News platform and then discuss its shortcomings and challenges. Then we propose our MOO based solution to this problem. 
\subsection{Article Ranking}
In this section we describe a simple Article Score as a linear additive function of user activities on each article (design variables). Higher user activities tends to correspond to better user-engagement. For each news article, $i$, we compute the article score as
  \begin{equation}
	Score_{i} = \alpha\cdot\textrm{Freshness} + 
     \beta\cdot\textrm{ Views} +
     \gamma\cdot\textrm{Likes} + 
     \phi\cdot\textrm{Comments} 
    \label{eq:score}
  \end{equation}
where $Freshness$ indicates how recent the article is; newer articles tend to be more interesting on a social news website, number of $Views$, number of $Likes$, and number of $Comments$ are user activity signals on the news article. $\alpha$, $\beta$, $\gamma$, and $\phi$ are the corresponding parameters of this scoring model. These parameters (or weights) are currently estimated using an optimization procedure called Grid Search. Once the values of the parameters are assigned, each article evaluates to an article score. This score is sorted to obtain a ranked list of articles.

The Grid Search uses a brute force search approach to uniformly sample the values of the parameters by finding the values of $\alpha$, $\beta$, $\gamma$, and $\phi$ that maximizes a particular scalar metric, like dwell-time. The data for this optimization task is generated on a small fraction of randomly chosen users, called the exploration bucket. User activities on articles and the user engagement metrics, like dwell-time, are recorded for each article for different choices of the four parameters we are trying to learn. The design assignment that generates the highest dwell-time is considered to be optimal and is assigned to  generate the scores used for the general (out of exploration bucket) user. The model is periodically updated to adapt to the changing dynamics in user behavior and traffic obligations. The reader can generalize this application to any optimization task which involves parameter (design variables) tuning of a function that evaluates to one or more metric (objectives) in an online optimization task.

While the current approach is simple and works reasonably well in practice, it involves searching over the entire space of parameters in a brute-force manner. The exploration bucket users are subjected to low-quality experience while we try to optimize these parameters. Searching over the entire space of possible parameters takes time and generally hits on user experience as well. We formulate this parameter searching problem as an optimization problem, and do a more principled and adaptive search of the parameter space. Based on the parts of the parameter space we have explored so far, our approach makes a decision on where to explore next and which areas need not be explored any further. 




\subsection{The MOO framework}
We first define the concept of \textit{non-dominance} that is closely related to the concept of Pareto optimality.

\noindent \textbf{Definition 2.} \textit{Non-Dominated and Dominated points: A vector of objective functions, $F(\vec{x}^*) \in C$, is non-dominated iff there does not exist another vector, $F(\vec{x}) \in C$, such that, $F(\vec{x}) \leq F(\vec{x}^*)$ with $f_i(\vec{x}) < f_i(\vec{x}^*)$ for at least one of the objectives. Otherwise, $F(\vec{x}^*)$ is said to be dominated.} \cite{marler2004survey}

The subtle difference between the concept of Pareto optimality (Definition 1) and non-dominance (Definition 2) is that Pareto optimality is defined in terms of the design space, while non-dominance is defined in terms of the objective space. For all practical purposes, this distinction is not important to us.

\cite{deb2002fast} propose a fast and efficient algorithm for non-dominated sorting (NDS) to sort solutions in the objective space and identify the set of non-dominated solutions. They use this NDS algorithm in their multi-objective evolutionary algorithm, NSGA-II, to identify the Pareto set in the solutions available in each generation. The computational complexity of this sorting approach is $O(mK^2)$, where $m$ is the number of objectives, and $K$ is size of the input to the NDS algorithm. In NSGA-II, this size is fixed to be equal to the user input parameter, population size.

\subsubsection{Grid Search (Baseline model)}
Grid Search is the current single-metric optimization technique used for our application. It is a simple brute force search over all possible values of the design variables. For each assignment of the design variables, it's objective evaluation is computed using the objective function. This objective function is a machine learning model, usually Gradient Boosted Decision Trees (GBDT) learned from the training data.

We extend this algorithm to handle multiple objectives. We do so by using the non-domination heuristic. We use NDS to identify the Pareto set from a set of different design variable assignments. We also extend the current approach to handle constraints by simply checking if each evaluation meets the constraints, before we sort it using NDS.

\begin{algorithm}[!ht]
\SetAlgoLined\DontPrintSemicolon
\KwIn{Training data ($D$), Design constraints ($S$), Objective functions ($F$), Increment percentage ($inc$)}
\KwOut{Approximated Pareto set, $Q$}
\SetKwFunction{Paretosort}{Non-Dominated-Sort}
\SetKwProg{gridsearch}{Grid-Search}{}{}
Initialize an empty list, $final \leftarrow \emptyset$.\;
Initialize a design vector $p$ with dimension $n$ assigning the minimum value for each design variable seen in the training data $D$.\;
Let $\Pi$ be a list of all possible permutations of values of $p$, where each variable takes discrete value increments, $inc\%$ inclusively between the minimum and maximum value of the design variable seen in $D$.\;
\ForEach{$\pi \in \Pi$}{
\If{$F(\pi)$ satisfies constraint $S$}{
Add $\pi$ to $final$\;}
}
$Q \leftarrow \text{Pareto-set in } \Paretosort{final}$\;
\Return{Q}
\caption{Grid Search (Baseline)}
\label{algo:gridsearch}
\end{algorithm}

Algorithm \ref{algo:gridsearch} describes our extension of the existing simple brute force algorithm, the Grid Search, to handle multiple objectives with constraints. The feasible solutions are sorted using NDS. The algorithm takes as input the training data, $D$ (from exploration bucket), Design constraints, $S$ (eg., traffic shaping requirements), Objective functions, $F$ (eg., GBDT model to predict Dwell time from an assignment of design variables.), and an increment, $inc$ to define the granularity of the different values explored for each design variables.

The time complexity of NDS is $O(mK^2)$ for $m$ objectives and $K$ solutions. In the worst case, Grid Search returns $v^{d}$ solutions, where $v$ is the number of different values generated from a granularity of $inc$. The overall complexity is $O(m v^{d})$. It can be clearly seen that it scales very poorly with the number of design variables, $d$.

\subsubsection{DO-NSGA-II}
In GA terminology, a collection of candidate solutions, $\vec{x}$, is called a \textit{population}. Typically, GAs initialize with random solutions as the initial population. GAs then use two operators to generate new solutions from existing ones: \textit{crossover} and \textit{mutation}. An iteration of generating new solutions from existing ones, using crossover and mutation operators, is called a \textit{generation}. In the \textit{crossover operator} two (or more) solutions, called \textit{parents}, are selected from the current population, their solutions are combined together to produce new solutions, called \textit{children}. Crossover ensures that elements of good solutions make it to the next generation promoting convergence. The \textit{mutation operator} introduces random changes to a solution. The mutation operator creates diversity in the solution set and also helps escape local optima. The crossover and mutation operator is sometimes preceded by the \textit{selection operator} that identifies parents from the population, which are promising candidates to approach convergence towards global optima, and passes them to the crossover and mutation operator. Each of the three operators: selection, crossover, and mutation, can be considered as separate modules in the GA.

We extend NSGA-II \cite{deb2002fast} (Non-dominated Sorting genetic Algorithm) from which we used the NDS algorithm to sort candidate solutions and identify the Pareto set in Grid Search. Other than the fast and efficient sorting algorithm NSGA-II is a popular choice in MOO literature. Another advantage of NSGA-II is that there exists a dynamic version to handle a dynamic objective function \cite{deb2007dynamic} which we can build on for our application. The authors test their approach with a numerical function and not in a data-driven online setting. We extend the NSGA-II framework to handle dynamic online data. We call this extension- Dynamic, Online NSGA-II or DO-NSGA-II.

\textit{Fitness function}: The fitness function of a GA is the heuristic used by the algorithm to evaluate the merit of a solution. DO-NSGA-II uses Pareto-dominance ranking as the fitness function. We define this heuristic in Definition 2. A set of solutions in a population, set A, in the GA is ranked better (lower) than another set of solutions, set B, if each individual solution in set A are non-dominated by each individual solution in set B.

\textit{Diversity}: An important advantage of using GA is the ability to control diversity of the population. DO-NSGA-II uses \textit{crowding distance} to prefer diverse solutions. A \textit{front} is a set of solutions in a population with the same Pareto-dominance rank. For each solution within a front, for each objective, $i$, the solutions within a front are first sorted with respect to that objective. The first and last solutions in the sorted list are given a distance score of infinity (to protect extreme solutions). For all the solutions in between, the score is computed by taking the normalized difference between the objective values of the solutions, before and after the current solution in the sorted list, Distance({$\vec{x}_i$}) $ \leftarrow \frac{f_j(\vec{x}_{i+1})-f_j(\vec{x}_{i-1})}{f_{j}^{max}-f_{j}^{min}}$. For a detailed review of different fitness functions and diversity metrics used in GA literature, refer to \cite{konak2006multi}.

\textit{Selection operator}: We use \textit{constraint dominance binary selection} \cite{deb2002fast} as our selection operator. This module handles the constraints in DO-NSGA-II and promotes diversity. It first randomly chooses two solutions from the current population. If one of the two is feasible, the feasible solution is selected to the next generation. If both are unfeasible, the solution with the smaller constraint violation is selected. If both solutions are feasible, the solution with the better Pareto rank is selected. If they belong to the same front, then the solution with the larger crowding distance is selected. The choice of this algorithm is particularly attractive to our problem where the algorithm adapts to dynamic changes in the constraints (changing traffic shaping targets).

\textit{Crossover operator}: We use \textit{Simulated Binary Crossover} (SBX) \cite{deb2001self} algorithm that is widely used in literature. We introduce two parameters to DO-NSGA-II that SBX depends upon: probability of crossover, $P_c$, and the spread factor of the polynomial distribution $\eta_c$. Smaller values of $\eta_c$ return children that are very different from the parents, while larger values return children similar to the parents.

\textit{Mutation operator}: We use \textit{Highly-Disruptive Polynomial Mutation} \cite{hamdan2012disruption} algorithm, shown to be effective at escaping the local optima on standard synthetic problems. In principle it is similar to SBX, and introduces similar parameters. the probability of mutation, $P_m$, and the spread factor for a polynomial distribution $\eta_m$. 

The purpose of the mutation operator in our application is to handle dynamics in the objective. Ideally we do not want a big loss in performance to a dynamic change of objective. We motivate our implementation using Hypermutation \cite{cobb1990investigation}. Here, we increase $P_m$ when a change is detected. We detect a change in DO-NSGA-II by re-evaluating all the existing solutions, if the objective value has changed from the previous generation, $P_m$ increases.

\begin{algorithm}[!htp]
\DontPrintSemicolon
\KwIn{Design constraints ($S$), Objective functions ($F$), Population size ($K$), Number of Generations ($E$), SBX parameters ($P_c$, $\eta_c$), Mutation parameters ($P_m$, $\eta_m$), Hypermutation period, ($epoch$)}
\SetKwFunction{crossover}{Crossover}
\SetKwFunction{mutation}{Mutation}
\SetKwFunction{Paretosort}{Non-Dominated-Sort}
\KwOut{Approximated Pareto set, Q}
Randomly initialize $K$ solutions, $G_0$. Sort the solutions, \Paretosort{$G_0$}.\;
$children \leftarrow $ \crossover{$\eta_c$, $G_0$} with a probability $P_c$ and \mutation{$\eta_m$, $G_0$} with a probability $P_m$.\;
\For{$t = 1\text{ to } E$}{
\If{objective change detected from $G_{t-1}$}{
Increase $P_m$ for $epoch$ generations.\;
}
$G_{t} \leftarrow G_{t-1} \cup children$\;
Sort the solutions, \Paretosort{$G_t$}\;
$G_t \leftarrow$ Choose top $K$ solutions from $G_t$ with the best Pareto rank. In case of a tie, choose solution with better crowding distance.\;
$children \leftarrow $ \crossover{$\eta_c$, $G_0$} with a probability $P_c$ and \mutation{$\eta_m$, $G_0$} with a probability $P_m$.\;
}
$Q \leftarrow\text{ Pareto-set in } \Paretosort{$G_{E}$}$\;
\Return{Q}
\caption{DO-NSGA-II}
\label{algo:donsga}
\end{algorithm}

The time-complexity of DO-NSGA-II remains $O(mK^2)$ for $m$ objectives and $K$ solutions.

\section{Experiments}

\subsection{Methodology}
We evaluate DO-NSGA-II and Grid Search using three experiments. For each of the experiment, we analyze an online dataset from a real-world problem of article ranking in a social news website as described in the Approach section. In the first experiment, we study the multi-objective optimization aspect of our framework. In the second experiment, we study the constraints handling capabilities of DO-NSGA-II and Grid Search. In the third and final experiment, we examine the dynamics handling capabilities of our framework. Grid Search cannot handle dynamic behavior without re-evaluating from the scratch.

For our experiments, we set the number of generations in DO-NSGA-II to $500$ and $500$ population size. The SBX parameters are $P_c = 0.9$, and $\eta_c = 15$, which are standard values used in literature. For mutation parameters, we set $P_m = 1/n$ and $\eta_m = 1$, as recommended by \cite{hamdan2012distribution}. We use our implemented hypermutation to increase $P_m = 1$ when a dynamic change in objective is detected.

\subsection{Evaluation}
In the introduction section, we mentioned that the two important properties of a good approximated Pareto set were convergence and diversity. \textit{Hypervolume} (Hv.) metric is the hypervolume (scalar value) occupied by some reference point and the approximated Pareto set treated as a minimization problem. This metric is commonly accepted since a large value indicates both better convergence and better diversity. For a solution set with $k$ solutions and $m$ objectives, we compute the hypervolume using an $O(k^{m-2}\log k)$ time and linear space complexity implementation as described in \cite{fonseca2006improved}. We perform min-max scaling of the objective space before computing the hypervolume to avoid bias in the metric from different scales of the dimensions of the objective space. Due to the min-max scaling, the worst result in the minimization problem cannot be greater than $(1.0, 1.0)$. So, we use a reference point value of $[2.0, 2.0]$.

We introduce a new metric called the \textit{Confidence Uncertainty} (CU), which is the volume (scalar value) occupied by the $90\%$ confidence intervals of each of the $m$ objectives. We estimate the $90\%$ confidence interval of each solution, across each objective, using quantile regression. We return CU as the product, of $m$ objectives, from the min-max normalized difference between the upper and lower bounds of the $90\%$ confidence interval.

\subsection{Datasets}
\subsubsection{Static data}
For the article ranking problem, we collect data over a week on a small fraction of randomly chosen users visiting the social news website. Each instance is a unique article on the website. There are $28922$ instances in the data. We collect user-behavior data on these articles. There are $4$ design variables: article freshness, total user views, likes, and comments. The $2$ objective variables of our interest are the total article clicks, and dwell-time (in milliseconds).

\subsubsection{Dynamic data}
The dynamics of our objectives are a function of time. The number of clicks or dwell time, any article receives depends upon the time of day. For example, it correlates to the work routine of a large subset of the users. Data was collected at $4$ different time intervals of the day: 1) Time step 1, [00:00, 6:00] hours ($50588$ instances), 2) Time step 2, [06:00, 12:00] hours ($51284$ instances), 3) Time step 3, [12:00, 18:00] hours ($72019$ instances), and 4) Time step 4, [18:00, 00:00] hours ($72705$ instances). We choose $4$ time steps spanning across a day to demonstrate the capabilities of DO-NSGA-II in handling dynamics. In practice, data can be collected over finer time-steps depending upon the application.

Article freshness is computed as the difference (in hours) between the article publish time of each article to the article with the latest publish time (smaller value indicates newer article). The other design variables (views, likes and comments) are count data. For each of the $4$ design variables and $2$ objectives ($x_i$ for instance $i$), we compute derived features by taking their $\log(x_i + 1\times10^{-5})$. We add $1\times10^{-5}$ to the count data as a smoothing factor to naturally handle the minimum value of zero when we take the log of the feature.

\subsection{Modeling objectives}
We split the training data into $80\%$ training set, $10\%$ test set, and $10\%$ validation set. We model the objectives using Gradient Boosted Decision Trees (GBDT) in the scikit-learn python package \cite{sklearn_api}. Optimal parameters for the GBDT are found using hyperopt \cite{bergstra2013making} while optimizing on the coefficient of determination ($R^2$) on the test set. The model performance on the validation data are reported as follows.

\subsubsection{Static data}
The Clicks model has an $R^2$ of $0.5323$ and a mean squared error (MSE) of $5.3213$. The Dwell time model has an $R^2$ of $0.56879$ and MSE of $4.6169$.

\subsubsection{Dynamic data}
In time steps $1$ through $4$, the four Clicks models have $R^2$ values of $(0.5223, 0.4701, 0.4629, 0.5059)$, and MSE values of $(5.8485, 6.4737, 6.2731, 5.9325)$. The four Dwell time models have $R^2$ values of $(0.5453, 0.5092, 0.5426, 0.5278)$, and MSE values of $(4.8183, 5.0986, 4.8743, 4.9541)$

\subsection{Experiment 1: Optimization performance}
\begin{figure}[!ht]
\centering
  \includegraphics[width=0.3\textwidth]{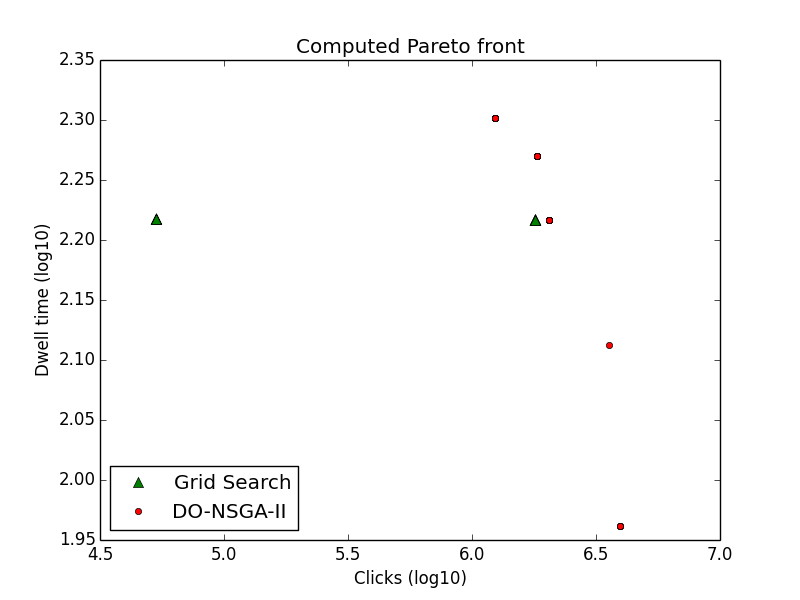}
  \caption{Comparing Pareto curve obtained by Grid Search (green triangles) and DO-NSGA-II (red circles)}
  \label{fig:exp1}
\end{figure}

In this experiment, we compare the quality of the Pareto curve returned by DO-NSGA-II and compare to Grid Search, the baseline algorithm. Figure \ref{fig:exp1} shows the Pareto curves we obtain from the two algorithms (shown as a maximization problem). We can immediately see that the solutions from DO-NSGA-II are better converged and spread than Grid Search.

\begin{table}[h]
\begin{center}
 \begin{tabular}{L{6em} C{5em} C{6em}}
 \hline
  & Grid Search & DO-NSGA-II \\
 \hline
  & & \\
 Total solutions & 45 & 699 \\ 
 Hv & 3.7161 & 3.7905 \\
 Average Hv & 3.6352 & 3.7256 \\
 \hline
\end{tabular}
\end{center}
\caption{Experimental results }
\label{table:exp1}  
\end{table}

The results in Table \ref{table:exp1} shows that Grid Search returns fewer solutions than DO-NSGA-II. The Figure \ref{fig:exp1} shows few data points because a lot of design variable assignments evaluate to a small number of different objective values. So, they are stacked in the figure. DO-NSGA-II achieves a much better overall Hypervolume. DO-NSGA-II also achieves a much better Average Hypervolume, the average  Hypervolume occupied by each individual solution in the approximated Pareto set. This measure shows that any arbitrary solution returned by DO-NSGA-II is likely to have a better Hypervolume than one from Grid Search.

\subsection{Experiment 2: Constrained MOO}
In this experiment, we observe the constraint handling capabilities of DO-NSGA-II and Grid Search. We set a target constraint on the Clicks objective, which could represent the number of clicks required in a traffic shaping problem for an article from a specific publisher to meet the contract requirements from that publisher. We test two scenarios— 1) Possible constraint, shows a reasonable number of clicks we need our article ranking model to achieve; 2) Impossible constraint, tests a constraint which would be unreasonable to achieve in light of observed training data.

\begin{figure}[!ht]
  \centering
  \subcaptionbox{Possible constraint\label{fig:exp2a}}{\includegraphics[width=0.23\textwidth]{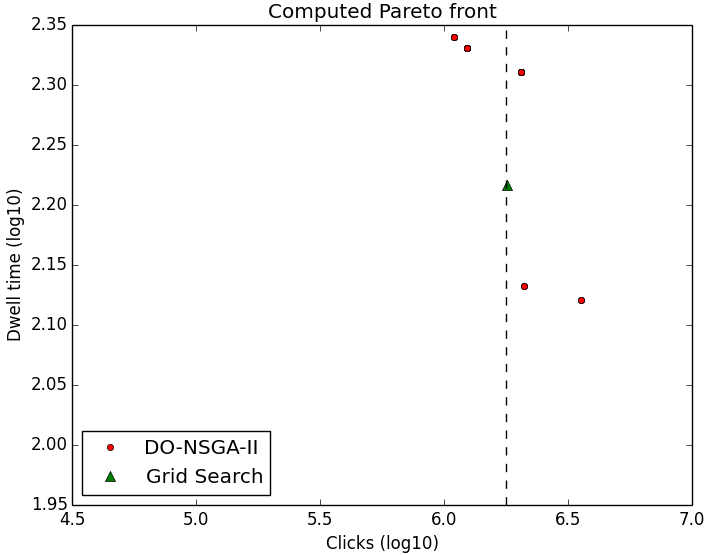}}\hspace{0.01em}%
  \subcaptionbox{Impossible constraint\label{fig:exp2b}}{\includegraphics[width=0.23\textwidth]{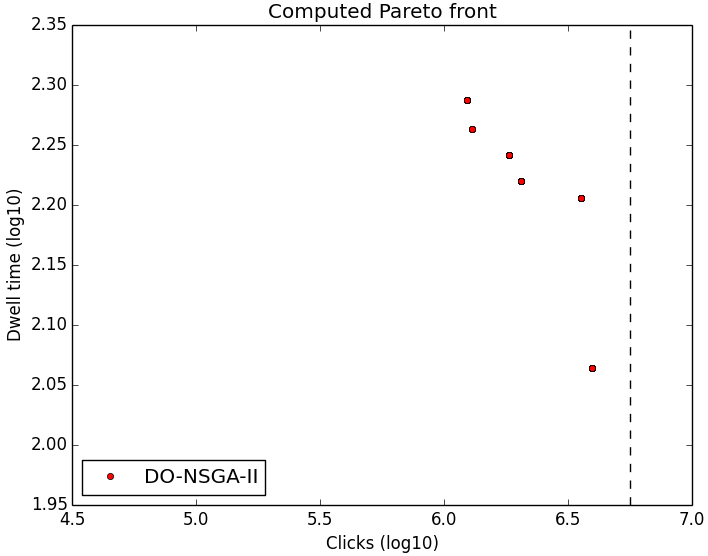}}
  \caption{DO-NSGA-II and Grid Search under constraints on Clicks objective.}
\end{figure}

Figure \ref{fig:exp2a} shows the result under achievable constraint of $\log_{10}(Clicks) > 6.25$. We observe that DO-NSGA-II, with its selection operator, adapts the search towards achieving the constraint. Grid Search cannot adapt and returns the same solution set as in Experiment 1. Fewer solutions meet the constraints. Figure \ref{fig:exp2a} shows the result under an impossible to achieve constraint of $\log_{10}(Clicks) > 6.75$. DO-NSGA-II still offers the best possible solution among the solutions that do not meet the constraints. For this reason, we chose this way of handling the constraints, organically, during search.

\subsection{Experiment 3: Dynamic optimization}
This experiment studies the dynamics handling capability of DO-NSGA-II. Grid Search cannot handle dynamics in the objective.
\begin{figure}[!ht]
  \centering
  \subcaptionbox{Pareto front over 4 time\\ steps.\label{fig:exp3a}}{\includegraphics[width=0.23\textwidth]{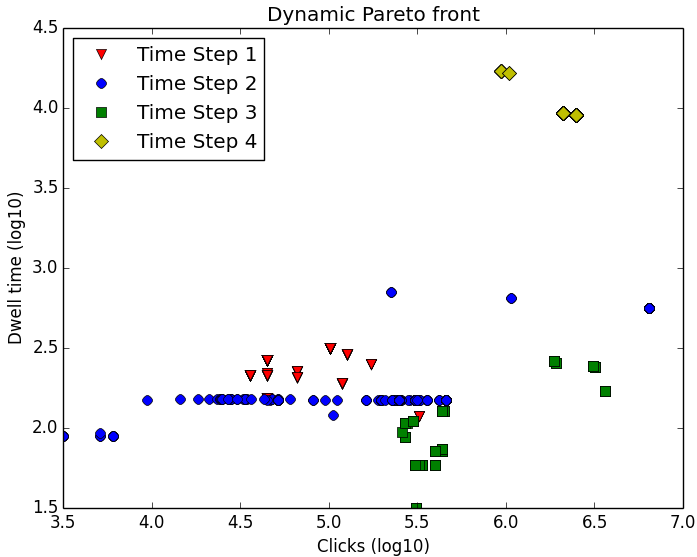}}\hspace{0.1em}%
  \subcaptionbox{Change in Hypervolume \\due to change in Pareto front.\label{fig:exp3b}}{\includegraphics[width=0.23\textwidth]{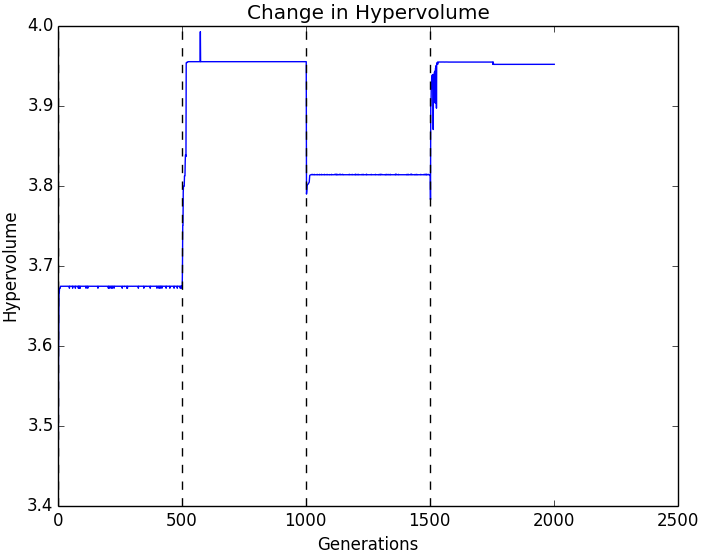}}
  \caption{DO-NSGA-II adapting under dynamics of different time steps.}
\end{figure}

Figure \ref{fig:exp3a} shows the changing pareto front through the four different time steps. The approximated pareto set in the different time steps is represented with different shapes and colors. Figure \ref{fig:exp3b} shows how the hypervolume changes as the algorithm adapts to the changing objective. We mark generations (0, 500, 1000, 1500) where the different time step models transition. Our design of hypermutation allows for a smooth transition between the different pareto fronts, allowing for reasonable solution at any given generation in an online setting.

\bibliographystyle{aaai}
\bibliography{bibliography.bib}

\end{document}